\documentclass[conference]{IEEEtran}
\IEEEoverridecommandlockouts
\usepackage{cite}
\usepackage{amsmath,amssymb,amsfonts}
\usepackage{algorithmic}
\usepackage{graphicx}
\usepackage{textcomp}
\usepackage{xcolor}
\usepackage{pgf}
\usepackage{booktabs}
\usepackage{hyperref}
\usepackage[linesnumbered, ruled, vlined]{algorithm2e}

\DeclareMathOperator*{\argmin}{arg\,min}
\DeclareMathOperator*{\argmax}{arg\,max}
\usepackage[hang,flushmargin]{footmisc}

\def\BibTeX{{\rm B\kern-.05em{\sc i\kern-.025em b}\kern-.08em
    T\kern-.1667em\lower.7ex\hbox{E}\kern-.125emX}}
\begin{document}

\title{Quantifying Predictive Uncertainty in Medical Image Analysis with Deep Kernel Learning}


\author{\IEEEauthorblockN{1\textsuperscript{st} Zhiliang Wu}
\IEEEauthorblockA{\textit{Siemens AG} \\
\textit{LMU Munich}\\
Munich, Germany\\
zhiliang.wu@siemens.com}
\and
\IEEEauthorblockN{2\textsuperscript{nd} Yinchong Yang}
\IEEEauthorblockA{\textit{Siemens AG} \\
Munich, Germany\\
yinchong.yang@siemens.com}
\and
\IEEEauthorblockN{3\textsuperscript{rd} Jindong Gu}
\IEEEauthorblockA{\textit{Siemens AG} \\
\textit{LMU Munich}\\
Munich, Germany\\
jindong.gu@siemens.com}
\and
\centering
\IEEEauthorblockN{4\textsuperscript{th} Volker Tresp}
\IEEEauthorblockA{\textit{Siemens AG} \\
\textit{LMU Munich}\\
Munich, Germany\\
volker.tresp@siemens.com}
}


\maketitle


\begin{abstract}
Deep neural networks are increasingly being used for the analysis of medical images.
However, most works neglect the uncertainty in the model's prediction. We propose an uncertainty-aware deep kernel learning model which permits the estimation of the uncertainty in the prediction by a pipeline of a Convolutional Neural Network and a sparse Gaussian Process. Furthermore, we adapt different pre-training methods to investigate their impacts on the proposed model.
We apply our approach to Bone Age Prediction and Lesion Localization. In most cases,  the proposed model shows better performance compared to common architectures. More importantly, our model expresses systematically higher confidence in more accurate predictions and less confidence in less accurate ones. Our model can also be used to detect challenging and controversial test samples. Compared to related methods such as Monte-Carlo Dropout, our approach derives the uncertainty information in a purely analytical fashion and is thus computationally more efficient.
\end{abstract}

\begin{IEEEkeywords}
uncertainty quantification, medical imaging, sparse Gaussian Process approximation, deep Convolutional Neural Networks, 
\end{IEEEkeywords}

\section{Introduction}

Various machine learning methods have been developed to support patient care to deal with the exploding amount of healthcare data\cite{tresp2016going, xiao2018opportunities}. An important example is medical imaging. In classical image analysis, the standard machine learning methods make predictions based on sophisticated handcrafted features extracted from the medical images. With the introduction of deep neural networks (DNNs), especially Convolutional Neural Networks (CNNs), manual feature engineering is replaced by self-organized supervised representation learning. 

Although DNNs now define the state-of-the-art in many applications, an often encountered  problem is that they fail to provide reasonable confidence estimates for their predictions\cite{nguyen2015deep}. In a classification task, this can result in over-confident predictions for misclassified samples \cite{guo2017calibration}. This observation has encouraged the development of various calibration methods such as temperature scaling\cite{zhang2020mix} and isotonic regression\cite{zadrozny2002transforming}. In a regression task, however, the modeling of input-dependent predictive uncertainty is rarely considered. Whereas in classification, a well-calibrated probabilistic prediction can sometimes be used to derive confidence values, in regression one needs to estimate confidence considering predictive distributions \cite{bishop1994mixture}.
\begin{figure}[!t]
    \centering
    \scalebox{0.85}{\includegraphics{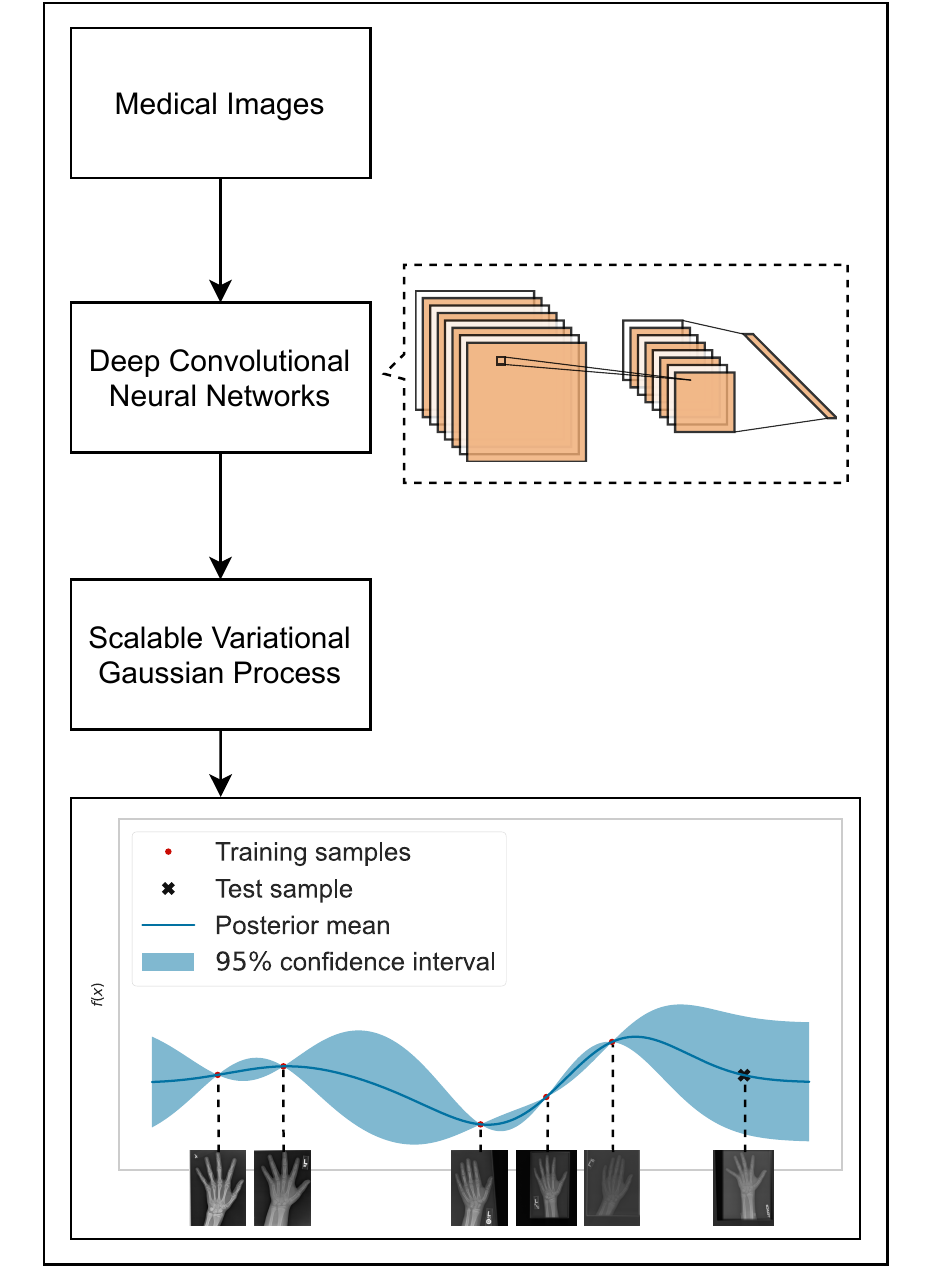}}
    \caption{An illustration of our proposed model: Combining a deep Convolutional Neural Network with the Scalable Variational Gaussian Process. Latent features of a medical image are extracted by the Convolutional Neural Network and then consumed by the Scalable Variational Gaussian Process. The proposed model outputs a predictive distribution of the Gaussian Process posterior, which can be interpreted as a mean estimate and a predictive variance. For instance, with a localized kernel the prediction of a rare test sample (the rightmost image) with few similar training samples demonstrates a larger variance.}
    \label{fig: dkl_draw}
\end{figure}

The problem of quantifying the predictive uncertainty in deep learning models limits their applicability to safety-critical domains such as healthcare \cite{kompa2021second, beam2018big}. In most cases, existing clinical decision support systems that rely on deep learning can only provide a point estimate, e.g., for a continuous severity score, progression-free survival time, or length-of-stay. 
Physicians, who are supposed to interpret the output of a DNN, face the challenge of not knowing how much they could trust the prediction. The goal of this paper is to provide a quantified uncertainty estimate for each prediction in a principled, mathematically sound manner. 
An unusually high uncertainty estimate would encourage the physician to investigate a case more closely since it is more likely to deviate from the ``normal" ones. A high uncertainty typically means that there are few similar cases in the training data. 

In a frequentist analysis of linear regression models, the predicted variance is derived from that of the model parameters; due to its high-dimensional nonlinear nature, this cannot easily be applied to DNNs. Currently, leading approaches to reason on the uncertainty in DNNs include MC dropout methods \cite{gal2016dropout, kendall2017uncertainties}, Bayesian neural networks \cite{bishop2006pattern, Kingma2014AutoEncodingVB} and deep ensembles \cite{lakshminarayanan2017simple}. These methods could become computationally expensive when large and deep neural networks are necessary. In this paper, we explore another direction to quantify predictive uncertainty with Gaussian Processes (GPs), a well-known class of Bayesian machine learning methods \cite{gpml2006}. A GP implicitly ties the predictive uncertainty with the similarity between samples, which does not model ensembles and can produce a predictive distribution with only one forward pass.

More specifically, when localized kernels are being used, e.g., Radial Basis Function (RBF) kernels or the more general Matérn kernels, 
a GP model would be confident in its mean estimate if there have been training samples observed in the ``neighborhood'' of the test input. Otherwise, the model would tend to output high variances for the predictions. Thus, for RBF-kernels, the Euclidean distance of inputs must be meaningful in the application, which often is not the case in high-dimensional problems, a typical example being raw images described by their pixel values.

Another known challenge in GP is the fact that computational complexity scales as  $\mathcal{O}(n^3)$ and storage complexity as $\mathcal{O}(n^2)$, where $n$ denotes the number of data samples \cite{gpml2006}. In recent years, significant progress has been made to address these scalability issues \cite{liu2020gaussian}, which has motivated work on combining DNNs with GPs, a.k.a. \textit{Deep Kernel Learning}\cite{wilson2016deep}. 
The pipeline architecture we propose in this paper is shown in  Fig.~\ref{fig: dkl_draw}, 
where we apply a state-of-the-art sparse GP on top of a CNN for predictions on medical images.  

Our contribution  can be summarized as follows:
\begin{itemize}
    \item We present a novel deep kernel learning model for regression on medical images based on the latest developments in sparse GPs and CNNs.  
    \item We enhance the proposed model by introducing different pre-training methods for the CNNs and initialization methods to optimize the inducing points in sparse GPs. 
    \item We apply the proposed model to the tasks of univariate bone age prediction and multivariate lesion localization, and provide a thorough comparison of different pre-training and initialization methods in terms of both point estimate and predictive uncertainty. 
\end{itemize}

\section{Related work}

\textbf{Deep Convolutional Neural Networks for Medical Image Analysis}
The analysis of medical images is arguably one of the areas where deep learning methods have been demonstrating the most promising performances, including diagnostics, dermatology, radiology, ophthalmology, and pathology\cite{esteva2019guide}. Deep learning-based solutions can offer physicians second opinions by, e.g., annotating the regions of interest. More specifically, CNN-based solutions have achieved physician-level accuracy, e.g., with CheXNet \cite{rajpurkar2017chexnet} for pneumonia detection, which is a 121-layer Dense Convolutional Network (DenseNet) \cite{huang2017densely} trained on the ChestX-ray 14 dataset\cite{wang2017chestx}. One factor limiting the progress is the relatively small size of labeled datasets available for specific clinical tasks when compared to large visual databases on nonmedical images, like the ImageNet dataset\cite{deng2009imagenet}. Therefore, methods like transfer learning are commonly used to take advantage of models trained on more or less unrelated datasets. However, many of these solutions focus only on improving the point estimate performance, ignoring the importance of the predictions' uncertainty. 
In this work, we focus on addressing the problem of providing meaningful uncertainty estimates. 

\textbf{Scalable Variational Gaussian Process with Neural Networks}
Efforts from earlier times include the Bayesian committe machine \cite{tresp2000bayesian}, Nyström methods \cite{williams2001using, williams2002observations},
the Fully Independent Training Conditional (FITC) Approximation \cite{snelson2005sparse}, and Variational Free Energy (VFE) \cite{titsias2009variational}.
Recently, \cite{hensman2013gaussian} proposed the Scalable Variational Gaussian Process (SVGP), which reduces the computational complexity to $\mathcal{O}(m^3)$, where $m$ denotes the number of \emph{inducing points} (more details see Sec.~\ref{sec: svgp}). In addition, SVGP enables the training with stochastic gradient descent (SGD)-based methods. 
Afterward, \cite{bradshaw2017adversarial} combined SVGP with DNNs for classification tasks. The approach is called Gaussian Process hybrid deep networks (GPDNN). 
\cite{wilson2016deep} combined neural networks with a KISS-GP covariance matrix, which takes advantage of a sparse matrix with inducing points lying on some grid structure \cite{wilson2015kernel}. The proposed method is called Deep Kernel Learning (DKL).
More recently, \cite{jankowiak2020parametric} proposed the  Parametric Predictive Gaussian Process (PPGP) regressor to improve the predictive variances in SVGP-based models, which shows promising performance in various applications. 
Inspired by the idea of DKL, we propose in this paper a model to train a state-of-the-art sparse GP model with deep CNNs in a more cohesive way.

\textbf{Pre-training Techniques for Deep Convolutional Neural networks}
Pre-training techniques have been developed for Deep Belief Networks \cite{hinton2006fast} and stacked auto-encoders\cite{bengio2007greedy}, where unsupervised pre-training was used for initialization, followed by supervised fine-tuning. Meanwhile, transfer learning techniques received increasing attention due to their ability to derive good representations for instances also from domains not considered in training\cite{yosinski2014transferable}. After the introduction of the ImageNet challenge \cite{deng2009imagenet}, it has been common practice to pre-train models on the ImageNet dataset as an initialization for other downstream tasks. More recently, self-supervised learning methods, e.g., contrastive learning\cite{chen2020simple}, have gained much attention as a powerful learning paradigm, which bridges the performance gap between supervised learning methods and unsupervised ones significantly. From a pre-training perspective, self-supervised learning can be considered as an example in deep metric learning (DML). The goal of DML is to map data to a latent space 
where data points with similar labels are located close together, and data with dissimilar labels are far apart \cite{musgrave2020metric}. 
In this paper, we adapt different pre-training methods under the setting of DKL.

\section{Method}
In this section, we provide a detailed introduction to our method. 
Our proposed model consists of two consecutive parts: a trainable feature extractor based on deep CNNs and an uncertainty-aware prediction model in the form of sparse GPs. The feature extractor is also commonly known as the \textit{backbone}, because it refers to the parts of the network excluding the final classification layers in, e.g., DenseNets \cite{huang2017densely} or ResNets\cite{he2016deep}. The output of such backbones, namely the feature maps, a.k.a. latent representations of the raw input, serves as input to the predictive GP regression. 
In the following, we first discuss how sparse GP models can scale to large datasets. 
Afterward, we introduce our initialization and pre-training techniques. Finally, we summarize the complete algorithm from the network initialization to the GP fine-tuning.

\textit{Notations}: We denote the training dataset as  $\{\boldsymbol{X}_i, {y}_i\}_{i=1}^n$, where $\boldsymbol{X}_i\in\mathbb{R}^{n_\text{H} \times n_\text{W}\times n_\text{C}}$ is an image of size $n_\text{H} \times n_\text{W}$ with $n_\text{C}$ color dimensions, ${y}_i\in \mathbb{R}$ is the target variable in a univariate regression task, and $n$ is the number of data samples. 
With the CNN backbones, we extract a latent representation from $\boldsymbol{X}_i$ and denote it as $\boldsymbol{h}_i$. 

\subsection{Scalable Variational Gaussian Processes as Output Layers}\label{sec: svgp}

A Gaussian process (GP) is a collection of random variables, any finite number of which have a joint zero-mean Gaussian distribution\cite{gpml2006}. Formally, if we denote all target variables $y_i$ in the column vector as $\boldsymbol{y}\in\mathbb{R}^n$ in a univariate regression problem, it follows
$$
\boldsymbol{y} \sim \mathcal{N}(\boldsymbol{0}, \boldsymbol{K} + \sigma_\text{obs}^{2} \boldsymbol{I}),
$$
where the covariance matrix $\boldsymbol{K}\in\mathbb{R}^{n\times n}$ is parametrized by the respective inputs as
$$
k_{ij} := k(\boldsymbol{h}_i, \boldsymbol{h}_j) 
$$
and $\sigma_\text{obs}^{2}$ is the noise variance.
Note that in a standard setup of GP, the input to the kernel function $k(\cdot, \cdot)$ is a pair of feature vectors. 
In the scope of our work, we feed the latent representations generated by CNN backbones to the kernel function. 

To find optimal hyperparameters in the kernel function (e.g., the scaling parameter in an RBF-kernel), the training of the GP involves maximizing the log marginal likelihood

$$
\mathcal{L}_\text{GP} = -\frac{1}{2} \boldsymbol{y}^{\top}\left(\boldsymbol{K}+\sigma_\text{obs}^{2} \boldsymbol{I}\right)^{-1} \boldsymbol{y}-\frac{1}{2} \log \left|\boldsymbol{K}+\sigma_\text{obs}^{2} \boldsymbol{I}\right|-\frac{n}{2} \log 2 \pi.
$$
Given a new input sample $\boldsymbol{h}_*$, the GP model provides a predictive distribution as 
\begin{align*}
    f_* &  \sim \mathcal{N}(\boldsymbol{k}^{\top}_{*}\left(\boldsymbol{K}+\sigma_\text{obs}^{2} \boldsymbol{I}\right)^{-1} \boldsymbol{y}, k_{**}-\boldsymbol{k}^{\top}_{*}\left(\boldsymbol{K}+\sigma_\text{obs}^{2} \boldsymbol{I}\right)^{-1} \boldsymbol{k}_{*}),
\end{align*}
where $\boldsymbol{k}_{*}=[k(\boldsymbol{h}_1, \boldsymbol{h}_*), \dots, k(\boldsymbol{h}_n, \boldsymbol{h}_*)]^\top \in \mathbb{R}^n$.

However, the complexity from the inverse operation of the large matrices in $\mathcal{L}_\text{GP}$ and $f_*$ hinders the application of GP models to large-scale datasets, which was the motivation for works on different approximation methods. 

The key idea of \cite{snelson2005sparse, titsias2009variational, hensman2013gaussian} is to learn a number of so-called inducing points by variational methods, which can be viewed as a learnable pseudo dataset $\{\boldsymbol{z}_i, u_i\}_{i=1}^m=:(\boldsymbol{Z}, \boldsymbol{u})$ to summarize the original large dataset, where $m\ll n$. The approximation follows these steps: 1) The original dataset is augmented with the inducing points; 2) Based on different assumptions, the log marginal likelihood $\log p(\boldsymbol{y})$ is approximated as a function only of inducing points; 3) Optimization is done by maximizing either the approximated log marginal likelihood \cite{snelson2005sparse}, or the lower bound of it, which is also known as the Evidence Lower BOund (ELBO) \cite{titsias2009variational, hensman2013gaussian}; 4) The predictions for new samples are based on the optimized inducing points instead of the original dataset. Among all approximation methods, SVGP turns out to be the most popular one, possibly thanks to its largely reduced computational and storage complexity as well as the natural integration of SGD-based methods\cite{hensman2013gaussian, hensman2015scalable}. Therefore, we take advantage of SVGP as one of the sparse GP models in this paper. 

In SVGP\cite{hensman2013gaussian}, a multivariate Normal distribution $\mathcal{N}(\boldsymbol{m}, \boldsymbol{S})$ is introduced to the variational distribution $q(\boldsymbol{u})$. \cite{jankowiak2020parametric} maximizes the ELBO 
\begin{equation}\label{loss: svgp}
\begin{aligned}
\mathcal{L}_\text{SVGP}=
&\sum_{i=1}^{n}\left\{\log \mathcal{N}\left(y_{i} \mid \mu_{\boldsymbol{f}}\left(\boldsymbol{h}_{i}\right), \sigma_{\text {obs }}^{2}\right)-\frac{\sigma^2_{\boldsymbol{f}}\left(\boldsymbol{h}_{i}\right)}{2 \sigma_{\text {obs }}^{2}}\right\} \\
&- \mathrm{KL}(q(\boldsymbol{u}) \| p(\boldsymbol{u} )),
\end{aligned}
\end{equation}
where we have the predictive mean $\mu_{\boldsymbol{f}}(\boldsymbol{h}_i)=\boldsymbol{k}_{i}^{\top}\boldsymbol{K}_{\boldsymbol{uu}}^{-1} \boldsymbol{m}$,
the predictive variance $\sigma^2_{\boldsymbol{f}}(\boldsymbol{h}_i) = k_{ii}- \boldsymbol{k}_{i}^{\top} \boldsymbol{K}_{\boldsymbol{ u u}}^{-1} \boldsymbol{k}_{i} + \boldsymbol{k}_{i}^{\top} \boldsymbol{K}_{\boldsymbol{uu}}^{-1} \boldsymbol{S} \boldsymbol{K}_{\boldsymbol{uu}}^{-1} \boldsymbol{k}_{i}$,
$p(\boldsymbol{u})=\mathcal{N}(\boldsymbol{0}, \boldsymbol{K}_{\boldsymbol{ u u}}), \boldsymbol{k}_{i}\in\mathbb{R}^m, \boldsymbol{K}_{\boldsymbol{ u u}}\in \mathbb{R}^{m\times m}$, and $\text{KL}(\cdot||\cdot)$ denotes the Kullback–Leibler divergence between two distributions. 
We use $\boldsymbol{\Theta}$ to denote all trainable parameters and adapt them with SGD-based methods, including $\boldsymbol{m}, \boldsymbol{S}$ for the variational distribution $q(\boldsymbol{u})$, inducing points $\boldsymbol{Z}, \boldsymbol{u}$ for the covariance matrices like $\boldsymbol{K}_{\boldsymbol{uu}}$ or $\boldsymbol{k}_{i}$, $\sigma_{\text{obs}}$ in the likelihood model and various hyperparameters in the kernel function, e.g., length scale in an RBF kernel.

\cite{jankowiak2020parametric} points out that the predictive distribution in SVGP tends to be dominated by the observational noise and underestimates the input-dependent uncertainty. 
As a solution, they proposed the PPGP Regressor, which takes advantage of the formulation of the predictive distributions in SVGP but restores the symmetry of the function variance $\mu_{\boldsymbol{f}}(\boldsymbol{h}_i)$ in the training objective through the maximum likelihood estimation (MLE) methods. Formally, the objective in PPGP is 
\begin{equation}\label{loss: ppgp}
\begin{aligned}
\mathcal{L}_\text{PPGP}
=&\sum_{i=1}^{n} \log \mathcal{N}\left(y_{i} \mid \mu_{\boldsymbol{f}}\left(\boldsymbol{h}_{i}\right), \sigma_{\mathrm{obs}}^{2}+\sigma^2_{\boldsymbol{f}}\left(\boldsymbol{h}_{i}\right)\right)\\
&- \mathrm{KL}(q(\boldsymbol{u}) \| p(\boldsymbol{u})).
\end{aligned}
\end{equation}
In our experiments, we report results of both SVGP and PPGP methods, which only differ in their respective ELBO objectives. 

Although the scalability problem in large datasets is nicely addressed in the SVGP-based models, the commonly used GP kernels, such as RBF and Matérn, cannot directly handle high dimensional data such as images. Therefore, there are many efforts to combine the inductive bias in neural networks and the non-parametric nature of GP-based models, including GPDNN \cite{bradshaw2017adversarial} and DKL \cite{wilson2016deep}. 
In \cite{bradshaw2017adversarial} and \cite{wilson2016deep}, the training is initiated with a standard neural network with a linear predictive model fit on the target variable. Afterward, the linear model is replaced with a GP to enable uncertainty-aware prediction. In our experiments with these approaches, we do not observe performance improvement in terms of point estimates. 
Therefore, we explore other pre-training methods that do not directly require the target variables, including Convolutional Autoencoders and Deep Metric Learning. 

\begin{figure}[!b]
    \centering
    \scalebox{0.85}{\includegraphics{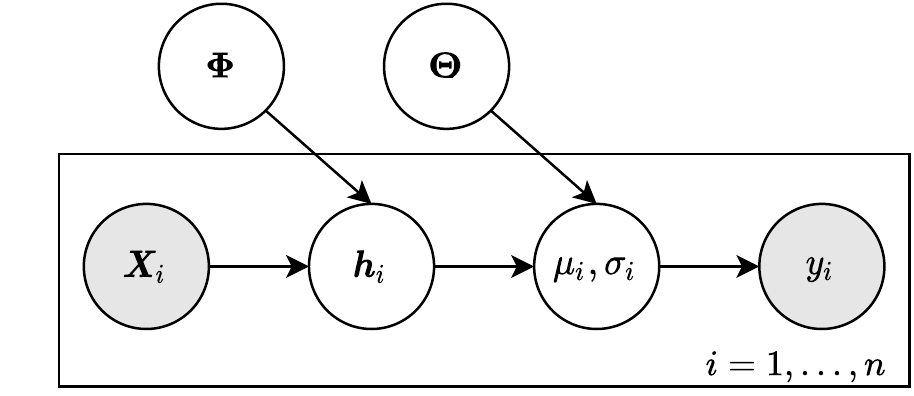}}
    \caption{Graphical model of deep kernel learning model in plate notation. Nodes are variables where shaded ones are observed, and non-shaded ones are latent variables. Plates indicate the repetition of the subgraph. }
    \label{fig: graph_dkl}
\end{figure}

Fig.~\ref{fig: graph_dkl} illustrates the basic idea of integrating DKL in our proposed model. The image sample $\boldsymbol{X}_i$ is embedded in some latent space defined by the backbone as $\boldsymbol{h}_i$. The SVGP-based model then consumes $\boldsymbol{h}_i$ as the input to produce the predictive distribution $\mathcal{N}(\mu_i, \sigma^2_i)$ for the target variable $y_i$. In other words, we propose to use SVGP-based models as output layers to replace the final linear layers found in common CNN architectures. The trainable parameters $\boldsymbol{\Phi}$ in the backbone and $\boldsymbol{\Theta}$ in the SVGP-based output layers are optimized together w.r.t. the ELBO objective.

Based on our observations, we realized that two modifications turn out to be critical for training the proposed model. First, we find it necessary for the model architecture to add one more linear layer after the backbone to \emph{further} reduce the dimension of the latent space. In ResNet18 and DenseNet121, the dimensions of the extracted latent spaces are 512 and 1024, respectively.
These dimensions prove to be too large for RBF kernel functions that rely on $l^2$ norm, presumably due to the fact that in high dimensional space, the Euclidean norm becomes irrelevant as a distance measure \cite{aggarwal2001surprising, beyer1999nearest}. With a thorough hyper-parameter search, we find that a dimension reduction to $50$ always shows a stable performance.

Second, the initialization of the inducing points plays an important role. If the inducing inputs are initialized from random vectors, the training never converges to meaningful results in our experiments. 
One explanation would be that with purely random initialization of inducing points, the covariance function value between each pair of samples is also random since it is defined via all inducing points (cf. Equation~(1) in \cite{titsias2009variational}). To this end, the GP has no chance of modeling the target based on these random distances with a multivariate Gaussian.
As a solution, we initialize inducing inputs by the latent representations produced by the backbone from the image samples. Formally, we initialize the inducing inputs by 
$$
\boldsymbol{z}^{\text{init}}_i = \boldsymbol{h}_i = f_{\boldsymbol{\Phi}}(\boldsymbol{X}_i),
$$
where the parameters $\boldsymbol{\Phi}$ in the backbone can be initialized from scratch, transferred from other models, or pre-trained in auxiliary tasks. Such a procedure is similar to using a subset of the dataset as the initial inducing inputs in a vanilla SVGP model, where raw features are fed to the model directly. 

So far, we focused on models for univariate regression problems. For the multivariate case, we propose to use the same backbone to generate the latent representations but feed it as input to multiple independent SVGP-based models. The number of involved SVGPs equals the dimension of the target variables.\footnote{This is also the configuration for the model with (multivariate) linear layers, which facilitates a fair comparison in experiments.} For the cases where there are correlations between the target variables, more advanced methods like Linear Model of Coregionalization (LMC) can be used \cite{alvarez2012kernels}, which we leave to our future work. 

\subsection{Pre-training Convolutional Neural Networks} \label{sec: cnn}
In our proposed architecture, the sparse GP model is defined in a latent space learned by the CNN backbone. The optimization task is correspondingly twofold: the GP is supposed to learn the parameters like the inducing points, and the CNN backbone should adapt its parameters to generate representative latent features. 
However, in the early phase of training, the CNN backbone may not have learned to extract representative features. In other words, the training samples could be mapped somewhat randomly in the latent space. This could pose a challenging task for the downstream GP model such that -- based on our observation of the experiments -- its parameters might converge to unfavorable values that are difficult to correct later on. 
To address this issue, we anticipate that the training quality could be improved by two strategies: 
\begin{itemize}
    \item Initialization with transfer learning in Sec. \ref{sec: transfer_learning} 
    \item Pre-training with auxiliary tasks: 
        \begin{itemize}
            \item Convolutional Autoencoder in Sec. \ref{sec: conv_autoencoder}
            \item Deep Metric Learning in Sec. \ref{sec: metric_learning}
        \end{itemize}
\end{itemize}

\subsubsection{Transfer learning} \label{sec: transfer_learning}
We regard transfer learning as reusing the knowledge from the models trained on different datasets. In the simplest case, we consider reusing CNN layers that have been trained on the classification task on the ImageNet dataset. It has been shown that in a CNN architecture, the early layers that are close to the input can learn to extract generic, low-level features that may apply across different types of image data\cite{yosinski2014transferable, sharif2014cnn}. These generic features typically include edges, basic patterns, and color gradients. We anticipate that such an initialization of the CNN would produce a latent space where the distance between samples better represents the distance in the original feature space, thus providing an advantageous starting point of learning the GP kernel. In the following, the pre-training methods are adapted to be used in settings either with or without transfer learning. With the proposed adaptations, we would be able to investigate the effectiveness of various components through experiments.

\subsubsection{Convolutional Autoencoder} \label{sec: conv_autoencoder}
The autoencoder (AE) is a well-known unsupervised representation learning method for dimensionality reduction. It consists of an encoder and a decoder. The encoder maps the inputs to some lower-dimensional latent space, whereas the decoder reconstructs the inputs from the latent representations, which ensures that the encoder has learned the most relevant features. Convolutional Autoencoders (CAEs) are a special case of AEs in that the convolutional filters are reused among different locations of the input to preserve the spatial locality \cite{masci2011stacked}. 

Normally, CAEs have several convolutional layers in the encoder and transposed convolutional layers in the decoder. To enable transfer learning in CAEs, we propose to use the CNN backbone as the encoder. And we construct a symmetric decoder using transposed convolutional layers.  In such a way, the decoder can have a similar model capacity to that of the encoder. Formally, after getting the latent representations $\boldsymbol{h}_i$ from the encoder $f_{\boldsymbol{\Phi}}(\cdot)$, we feed it into the decoder $g_{\boldsymbol{\Psi}}(\cdot)$ to reconstruct the original images
\begin{align*}
    g_{\boldsymbol{\Psi}}:  \mathbb{R}^h & \to  \mathbb{R}^{n_\text{H} \times n_\text{W}\times n_\text{C}} \\
    \boldsymbol{h}_i & \mapsto g_{\boldsymbol{\Psi}}(\boldsymbol{h}_i)=:\hat{\boldsymbol{X}}_i,
\end{align*}
where $\boldsymbol{\Psi}$ denotes the trainable parameters in the decoder network. The parameters of the encoder $\boldsymbol{\Phi}$ can be initialized from scratch or from models using transfer learning. The training of the CAE involves minimizing the Mean Squared Error (MSE) between the original image sample $\boldsymbol{X}_i$ and the reconstructed image $\hat{\boldsymbol{X}}_i$. Formally, the loss function is defined as 
\begin{equation}\label{loss: cae}
    \mathcal{J}_\text{CAE} = \frac{1}{n}\sum_{i=1}^n \|\boldsymbol{X}_i - \hat{\boldsymbol{X}}_i\|_2^2,
\end{equation}
where $\|\cdot\|_2$ denotes the Euclidean norm.

\subsubsection{Deep Metric Learning} \label{sec: metric_learning}
Taking advantage of the structure of twin neural networks (replications of the same NN), CNN backbones are applied in DML for learning the latent representations so that samples with similar labels would be mapped closer to each other in the latent space. The latent representations from the trained backbones turn out to be effective for tasks like face verification \cite{taigman2014deepface, schroff2015facenet} or person re-identification \cite{hermans2017defense}.

Given an image sample $\boldsymbol{X}_i$ from the dataset, the backbone defines a function $ f_{\boldsymbol{\Phi}}(\cdot)$ to embed it in some latent space. Formally, we have 
\begin{align*}
    f_{\boldsymbol{\Phi}}: \mathbb{R}^{n_\text{H} \times n_\text{W}\times n_\text{C}} & \to \mathbb{R}^h \\
    \boldsymbol{X}_i & \mapsto f_{\boldsymbol{\Phi}}(\boldsymbol{X}_i)=:\boldsymbol{h}_i,
\end{align*}
where $\boldsymbol{\Phi}$ denotes the trainable parameters in the network and $h$ is the dimension of the latent space. The trainable parameters $\boldsymbol{\Phi}$ can be either initialized from scratch or be transferred from models trained on other large-scaled datasets, e.g., the ImageNet dataset. 

With the class label information, a triplet is defined to consist of an anchor sample $\boldsymbol{X}^\text{A}_i$, a positive sample $\boldsymbol{X}^\text{P}_i$, and a negative sample $\boldsymbol{X}^\text{N}_i$, where the anchor is of the same class as the positive and the negative is not. However, in a regression problem, the target variables cannot define the triplets directly since they are continuous values. To mitigate this, we propose to categorize the target variables into classes to generate triplets in DML. It can also be viewed as a coarse pre-training step before the final fine-tuning from a learning perspective. Concretely speaking, we categorize target variables according to their binning in the histogram for univariate regression tasks and apply K-means clustering to find a class label for the target variables in multivariate regression tasks. 

During training, minimizing the triplet margin loss makes the anchor-positive distance smaller than the anchor-negative distances by a certain margin \cite{schroff2015facenet}. Formally, the loss function is defined as 
\begin{equation}\label{loss: triplet}
    \mathcal{J}_\text{triplet}=\sum_{i=1}^n \big[\text{d}(\boldsymbol{h}^\text{A}_i, \boldsymbol{h}^\text{P}_i) -  \text{d}(\boldsymbol{h}^\text{A}_i, \boldsymbol{h}^\text{N}_i) + \alpha]_+,
\end{equation}
where $\text{d}(\cdot, \cdot)$ is a distance metric, e.g., the Euclidean distance, $\alpha$ is the value of a pre-defined margin, and $[\cdot]_+$ only takes the positive part of the variable. Also, triplet selection is an important step to get fast convergence of the training since the network only gets gradients from the triplets having positive values in Equation~(\ref{loss: triplet}). 
Within each mini-batch, we pick negative samples whose distance to the anchor is larger than the anchor-positive distance (within a margin of $\alpha$). That means we have
$$
0 < \text{d}(\boldsymbol{h}^\text{A}_i, \boldsymbol{h}^\text{N}_i) - \text{d}(\boldsymbol{h}^\text{A}_i, \boldsymbol{h}^\text{P}_i) < \alpha,
$$
which are regarded as \textit{semi-hard} examples in \cite{schroff2015facenet}. 

To find an appropriate number of epochs for the pre-training, we take advantage of early stopping methods, 
which terminate the training automatically by monitoring specific metrics derived from the validation set. Here, we use the metric Mean Average Precision at R (MAP@R), a more informative evaluation metric since it combines the ideas of Mean Average Precision and R-precision \cite{musgrave2020metric}.

We argue that the training objective of the DML agrees with the paradigm of a GP regression using localized kernels, which is to interconnect the similarity of data samples in the target space to the similarity in their input spaces. Since GP cannot directly operate in the raw pixel space, a mapping function that preserves the similarity from the target space to the latent space would provide the GP with an ideal input space.   

\subsection{End-to-end Fine-tuning Deep Kernel Learning}
In this section, we elaborate our proposed method in Algorithm~\ref{alg: ftdkl} by inversely joining the modules that have been introduced in the last two sections.
\begin{algorithm}[!htb]
	\SetKwInput{KwIn}{Input}
	\SetKwInput{KwOut}{Output}
    \SetAlgoLined
\KwIn{An image dataset of the form $\{\boldsymbol{X}_i, y_i\}_{i=1}^n$.}
\KwOut{A fine-tuned DKL model for regression 
}
\If{Transfer is True} 
{
$\boldsymbol{\Phi}\leftarrow\boldsymbol{\Phi}^{\text{ImageNet}}$ \label{alg: transfer}
}
\Switch{Pre-training is DML}
{Generate triplets $\{\boldsymbol{X}_i^\text{A}, \boldsymbol{X}_i^\text{P}, \boldsymbol{X}_i^\text{N}\}$  \label{alg: triplet_generation}\\ 
$\boldsymbol{\Phi}\leftarrow \argmin_{\boldsymbol{\Phi}}\mathcal{J}_{\text{triplet}}(\boldsymbol{\Phi}) $\label{alg: triplet_loss}
}
\Switch{Pre-training is CAE }
{$\boldsymbol{\Phi}, \boldsymbol{\Psi}\leftarrow \argmin_{\boldsymbol{\Phi}, \boldsymbol{\Psi}}\mathcal{J}_{\text{CAE}}(\boldsymbol{\Phi}, \boldsymbol{\Psi}) $ \label{alg: cae}
}
Initialize the inducing points $\{\boldsymbol{Z}| \boldsymbol{z}^{\text{init}}_i = f_{\boldsymbol{\Phi}}(\boldsymbol{X}_i) \}$ \label{alg: init_inducing} \\
$\boldsymbol{\Phi}, \boldsymbol{\Theta} \leftarrow \argmax_{\boldsymbol{\Phi}, \boldsymbol{\Theta}} \mathcal{L}_{\text{PPGP}}(\boldsymbol{\Phi}, \boldsymbol{\Theta})$\label{alg: fine_ppgp}\\
\KwRet{$\boldsymbol{\Phi}, \boldsymbol{\Theta}$}
\caption{\textbf{Fine-tuning Deep Kernel Learning}} 
\label{alg: ftdkl}
\end{algorithm}

Depending on whether we want to reuse the model trained on the ImageNet dataset, we will initialize the parameters in the backbones from the transferred model or from scratch (line~\ref{alg: transfer}). If we use DML as pre-training, we first categorize the target variables to generate triplets with the class information (line~\ref{alg: triplet_generation}). Then the backbones are trained with triplet margin loss in Equation~(\ref{loss: triplet}) (line~\ref{alg: triplet_loss}). On the other hand, if we use CAE as pre-training, the parameters in the encoder and decoder are trained jointly against the CAE loss in Equation~(\ref{loss: cae}) (line~\ref{alg: cae}), where the encoder will be used as the backbone in later steps. After the pre-training, the backbone is used to initialize the inducing points with a subset of the image samples (line~\ref{alg: init_inducing}). Therefore, it is worth highlighting that the pre-training step affects the parameters in the neural network and the parameters in the SVGP-based output layer. Finally, the fine-tuning step is done w.r.t. the respective ELBO objective, where we use the PPGP objective in Equation~(\ref{loss: ppgp}) as an example (line~\ref{alg: fine_ppgp}).

\section{Experiments}\label{sec: exp}
\subsection{Datasets and Implementation Details}
\begin{figure}[t]
    \centering
    \scalebox{0.9}{\includegraphics{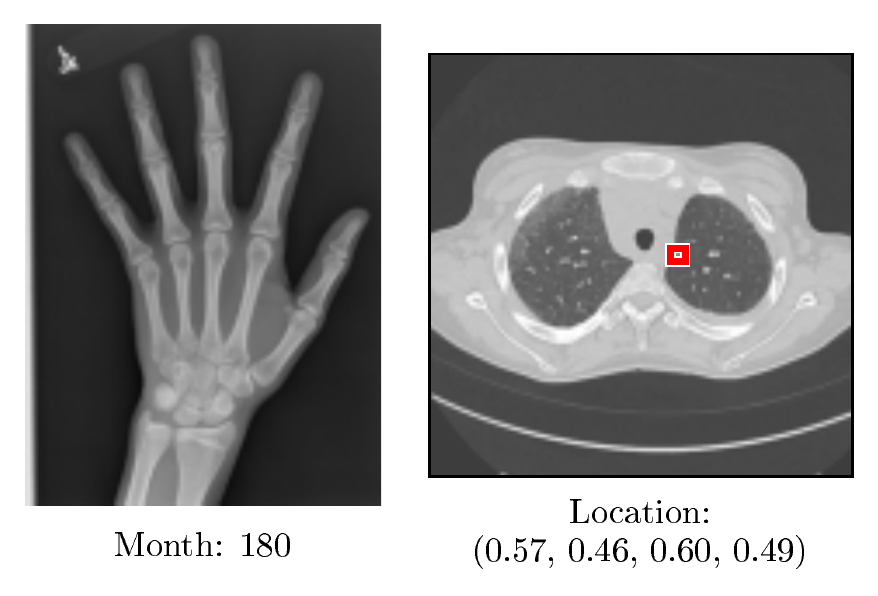}}
    \caption{An example for the task of Bone Age Prediction (left) and Lesion Localization (right). }
    \label{fig: task_viz}
\end{figure}

We have conducted experiments with two different datasets to validate our proposed methods. As an example for univariate regression tasks, we included the Bone Age Prediction (BAP) task from the Radiological Society of North America  Pediatric Bone Age Machine Learning Challenge \cite{halabi2019rsna, larson2018performance}. In this dataset, there are $14,236$ hand radiographs, where the target variable is defined as the bone age of pediatric patients under five years old.
In addition, we included the lesion localization (LL) task from the DeepLesion dataset \cite{yan2018deeplesion} as an example for the multivariate regression problem. In the original dataset, there are $32,120$ axial computed tomography (CT) slices from $4,427$ unique patients. Together with the tag information from LesaNet \cite{yan2019holistic}, we retrieve $7,310$ slices for the lesion type of lung, where the task is to localize the lesion in a given CT image. The target is in the format of $(x_{1n}, y_{1n}, x_{2n}, y_{2n})$, where $x_{1n}, y_{1n}, x_{2n}, y_{2n}$ denote the normalized $x$-top-left, $y$-top-left, $x$-bottom-right, and $y$-bottom-right, respectively. Fig.~\ref{fig: task_viz} shows examples for the task of BAP and LL. 

The CNN-related models are built using the \textit{PyTorch} package\cite{NEURIPS2019_9015}, where the models with GP methods are implemented with the help of the \textit{GPyTorch} package \cite{gardner2018gpytorch}. We conducted cross-validations (CV) for both tasks with $90\%$ samples extracted in the datasets, where hyperparameters are tuned according to the performance on the validation set. The remaining unseen $10\%$ samples constitute the test set, from which the results reported in the following sections are computed. Common data augmentations, including random crop, rotate, and horizontal flip, are applied. 
Due to the relatively small size of the datasets, the backbone of ResNet18 and DenseNet121 are chosen in all experiments. Related scripts\footnote{Related scripts see \url{https://github.com/ZhiliangWu/mDKL}.} of the work will be published to improve the reproducibility.

\subsection{Evaluation Approaches and Baselines}

Due to the probabilistic nature of our proposed model, we considered two lines of evaluation approaches in the experiments: the performance of point estimates and the evaluation of predictive variances. We used the well-known Root Mean Squared Error (RMSE) to reflect the prediction performance for the former, where only the mean predictions of the proposed model are involved in the evaluation. For the latter, we included a novel method to validate the meaningfulness of the predictive uncertainty, namely a quantile performance (QP) plot. Intuitively speaking, a good uncertainty-aware model should demonstrate better performance together with higher confidence in its predictions and vice versa. Any uncertainty-aware regression model that produces point estimates and predictive variances can be evaluated against this criterion. 

Given an uncertainty-aware model $f(\cdot)$ and its predictive distribution $f_i \sim \mathcal{N}(\mu_i, \sigma_i^2)$, 
we first sort all predictive variances in an ascending order $\{\sigma_i^2~|~\sigma_i^2 \leq \sigma_{i+1}^2, i \in \{1,\dots, n\} \}$ and then compute $K$ quantiles denoted as $q_1, \dots, q_K$.

Second, we compute the RMSE evaluation\footnote{This could be any metrics for evaluating point estimates.} of the subset of predicted point estimates $\hat{y}_i:=\mu_i$, whose paired variances $\sigma_i^2$ are smaller than or equal to each of the $k$-th quantile values of $k \in \{1, \dots, K\}$: 
$$
\text{Performance}_k := \text{RMSE}(\{(y_i, \hat{y}_i)|\forall \sigma_i^2 \leq k\text{-quantile} \}),
$$
where $(y_i, \hat{y}_i)$ denotes the evaluation pair. By plotting the performance value on the $y$-axis against the corresponding quantile value $k$ on the $x$-axis, a monotonically increasing line is expected. 

To study the point estimate performance of the SVGP-based output layers, models having the same backbone but with a linear layer, which is optimized directly w.r.t. MSE, are included as baselines. Besides, when investigating the effects of pre-training between various representation learning methods, the models without any pre-training serve naturally as baselines. For the performance of predictive variances, we included MC Dropout \cite{gal2016dropout}, a popular method for augmenting uncertainty in the neural networks, as a baseline. In MC Dropout, a dropout layer \cite{srivastava2014dropout} is added before each layer in the network. In our experiments, we used the default dropout setting for DenseNet121 with a dropout rate of 0.2 during training and testing, whereas dropout layers with the same dropout rate are added after each of the four layers of residual blocks in ResNet18. The predicted value and predictive variances are computed by performing 50 stochastic forward passes through the network as suggested in \cite{srivastava2014dropout}.  

\subsection{Evaluation on the Bone Age Prediction}

\subsubsection{Results on Point Estimates}

Tab.~\ref{tab: ba_dense} demonstrates the performance of the proposed method with DenseNets121 on the univariate regression task, Bone Age Prediction. Our proposed models with SVGP-based output layers deliver competitive or, in most cases, even better performances compared to common architectures with linear layers. Overall, the proposed model with SVGP output layers using transfer learning and pre-trained with DML demonstrates the best performance. In addition, significantly superior performance is found on models using the parameters transferred from models trained on the ImageNet dataset, which holds under all pre-training variants. Comparing the models using transfer learning w.r.t. different pre-training methods (upper part in Tab.~\ref{tab: ba_dense}), we see an improvement when the model is first pre-trained with DML, whereas the pre-training with CAE does not improve the performance. However, for the models without transfer learning (lower part in Tab.~\ref{tab: ba_dense}), both DML and CAE enhance the performance of the models, whereas CAE shows a better pre-training performance than DML. As a reference, \cite{larson2018performance} reports $10.44$ and $7.8$ as RMSE values on $200$ test samples from human reviewers and model predictions, respectively.
\begin{table}[!tb]
\begin{minipage}{0.5\textwidth}
\centering
\caption{Bone Age Prediction with DenseNet121}
\label{tab: ba_dense}
\scriptsize
\begin{tabular}{p{0.5cm}cccc}
\hline
\begin{tabular}[c]{@{}c@{}}Output \\ Layer\end{tabular} & \begin{tabular}[c]{@{}c@{}}Transfer \\ Learning\end{tabular} & \begin{tabular}[c]{@{}c@{}}RMSE\\ (No pre-training)\end{tabular} & \begin{tabular}[c]{@{}c@{}}RMSE\\ (DML)\end{tabular} & \begin{tabular}[c]{@{}c@{}}RMSE\\ (CAE)\end{tabular} \\ \hline
Linear\footnote{Common architectures.}    & Yes                                                          & $12.118 \pm 0.277$                                              & $11.667 \pm 0.231$                                      & $14.076 \pm 0.281$                                   \\
SVGP\footnote{With our proposed model.}                   & Yes                                                          & $11.697 \pm 0.102$                                              & $\mathbf{11.440 \pm 0.132}$                             & $\mathbf{13.536 \pm 0.279}$                          \\
PPGP$^\dagger$                                      & Yes                                                          & $\mathbf{11.679 \pm 0.061}$                                     & $11.529\pm 0.089$                                       & $13.694\pm 0.274$                                    \\ \hline \hline
Linear$^*$                                                  & No                                                           & $19.934 \pm 0.246$                                              & $\mathbf{15.805\pm0.157}$                               & $15.340\pm0.390$                                     \\
SVGP$^\dagger$                                      & No                                                           & $\mathbf{17.723 \pm 0.298}$                                     & $15.832\pm 0.284$                                       & $\mathbf{15.323\pm 0.411}$                           \\
PPGP$^\dagger$                                    & No                                                           & $18.341 \pm 0.234$                                              & $16.084\pm0.336$                                        & $15.752\pm0.352$                                     \\ \hline
\end{tabular}
\end{minipage}
\end{table}
\subsubsection{Results on Predictive Variances}
\begin{figure}[tb]
    \centering
    \scalebox{0.9}{\input{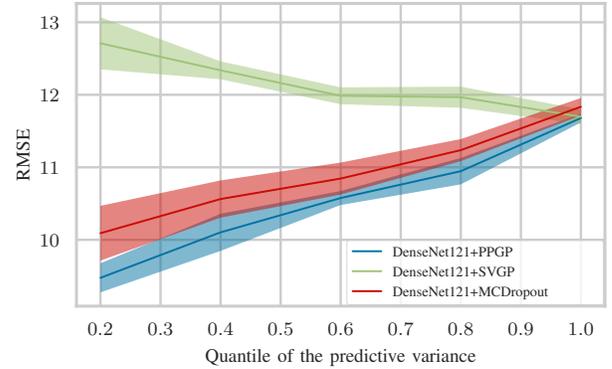}}
    \caption{Quantile Performance for the Bone Age Prediction with DenseNet121}
    \label{fig: qp_ba_dense}
\end{figure}
\begin{figure}[tb]
    \centering
    \input{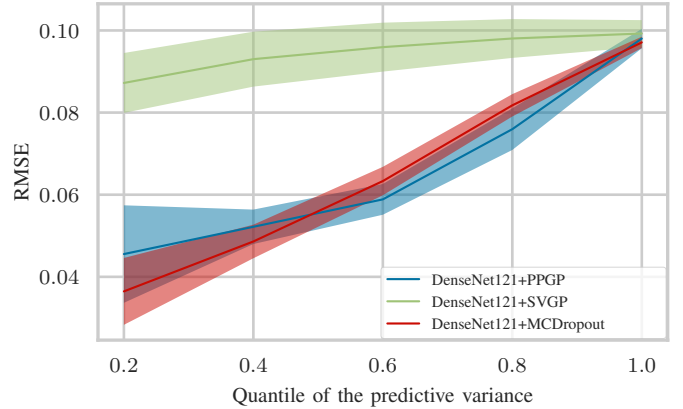}
    \caption{Quantile Performance for the Lesion Localization with DenseNet121}
    \label{fig: qp_dl_dense}
\end{figure}

We include the models with transfer learning but without any pre-training for the QP plot in Fig.~\ref{fig: qp_dl_dense}, where solid lines and error bars denote the means and standard deviations across different CV splits. A clear, monotonically increasing trend is observed in our proposed models with PPGP output layers and the models with MC Dropout. The line of PPGP is located underneath the MC Dropout, indicating better performance. In contrast, the models with SVGP output layers show a monotonically decreasing trend w.r.t. the quantile of the predictive variance. The models with PPGP output layers have an RMSE of $9.476 \pm 0.200$ (predictive variances at $q=20\%$) for the samples they are more confident with, which is a relatively large improvement compared to the values reported in Tab.~\ref{tab: ba_dense}. 

\subsubsection{Discussion}
The superior performance of models with SVGP-based output layers is expected since the ELBO objective is a proxy for the log marginal likelihood objective, which is a generalization to MSE in linear regression. The improvements based on transfer learning conform to its popularity in the computer vision community, which validates the hypothesis that reusing the knowledge from large-scale datasets could help solve a new problem even with domain shift. For pre-training with DML, the models are first trained to embed samples with similar targets into nearby regions, which would be a helpful initialization for the inducing points and non-convex optimization in neural networks. Therefore, we observe a positive contribution from DML to the model performance. For the pre-training with CAE, the goal is to learn a \textit{compressed} representation, which will be recovered in another decoder network. From the experimental results, such unsupervised representation learning would improve the performance if the model is trained from scratch but may deteriorate the knowledge transferred from large-scale datasets. It indicates a possibly higher correlation of the current task to the ImageNet classification than the compression task from CAE, which could be attributed to the large number of training samples in the ImageNet dataset. It is also worth mentioning that we also conducted the same experiments with the ResNet18 backbone, where similar results are observed. More details can be found in Appendix~\ref{app: point_resnet} and \ref{app: pred_resnet}.

What makes our proposed method appealing lies in the probabilistic nature of its prediction. The principled predictive variance from PPGP output layers is expected since 1) the inducing points technique facilitates explicit modeling of uncertainty, 2) the symmetric treatment of the predictive variance is restored in the training phase compared with SVGP. However, the good performance from MC Dropout also comes with a considerable computational cost. Roughly speaking, the inference time would be $t$ times as much as our proposed model, where $t$ is the number of stochastic forward passes for the inference. With $t=50$, our experiment with 1424 test samples requires an inference time of $1777.04$ seconds (almost half an hour) for the MC Dropout method, whereas our SVGP-based approach takes only $42.90$ seconds. With more samples or more advanced backbone structures, the time cost will be more expensive for the MC Dropout method. These observations indeed motivate the application of our proposed models with PPGP output layers when meaningful predictive variances and low time complexity come to a higher priority in a real-world system.

\subsection{Evaluation on the Lesion Localization}
\subsubsection{Results on Point Estimates}
Tab.~\ref{tab: dl_dense} shows the performance of our proposed method with DenseNet121 on the multivariate regression task, Lesion Localization. Similar to the task of BAP, our proposed models with SVGP-based output layers have mostly better performance than the common architectures with linear layers, and models with transfer learning outperform the ones without it by a large margin. On the whole, the proposed model with PPGP output layers demonstrates the best results under all settings. The difference lies in the performance with pre-training methods. Both pre-training methods only improve the performance in the settings without transfer learning. 

\subsubsection{Results on Predictive Variances}
Due to the multivariate setting in this task, the mean of the predictive variances of different target variables is first computed before quantifying the predictive variances. Like the BAP task, a monotonically increasing trend is observed in models with PPGP and MC Dropout. The line of PPGP overlaps mostly with the one with MC Dropout, indicating relatively similar performance. In contrast, models with SVGP output layers deliver an almost flat trend w.r.t. the quantiles of the predictive variance. The models with the PPGP output layers have RMSE of $0.046\pm 0.012$ for the evaluation pairs they are more confident with (predictive variance at $q=20\%$), which is less than one-half of the ones reported in Tab.~\ref{tab: dl_dense}. 
\begin{table}[!tb]
\begin{minipage}{0.5\textwidth}
\centering
\caption{Lesion Localization with DenseNet121}
\label{tab: dl_dense}
\begin{tabular}{p{0.5cm}cccc}
\hline
\begin{tabular}[c]{@{}c@{}}Output \\ Layer\end{tabular} & \begin{tabular}[c]{@{}c@{}}Transfer \\ Learning\end{tabular} & \begin{tabular}[c]{@{}c@{}}RMSE\\ (No pre-training)\end{tabular} & \begin{tabular}[c]{@{}c@{}}RMSE\\ (Metric)\end{tabular} & \begin{tabular}[c]{@{}c@{}}RMSE\\ (CAE)\end{tabular} \\ \hline
Linear\footnote{Common architectures.}                       & Yes                                                          & $0.102 \pm 0.002$                                               & $0.101 \pm 0.002$                                       & $0.102 \pm 0.003$                                    \\
SVGP\footnote{With our proposed model.}                      & Yes                                                          & $0.099 \pm 0.003$                                               & $0.101 \pm 0.003$                                       & $0.104 \pm 0.004$                                    \\
PPGP$^\dagger$                                           & Yes                                                          & $\mathbf{0.098 \pm 0.002}$                                      & $\mathbf{0.098 \pm 0.002}$                              & $\mathbf{0.099 \pm 0.002}$                           \\\hline \hline
Linear$^*$                                         & No                                                           & $0.116 \pm 0.001$                                               & $0.114 \pm 0.003$                                       & $0.114 \pm 0.002$                                    \\
SVGP$^\dagger$                                       & No                                                           & $0.118 \pm 0.002$                                               & $0.114 \pm 0.003$                                       & $0.112 \pm 0.003$                                    \\
PPGP$^\dagger$                                      & No                                                           & $\mathbf{0.115 \pm 0.002}$                                      & $\mathbf{0.111 \pm 0.005}$                              & $\mathbf{0.110 \pm 0.002}$                           \\ \hline
\end{tabular}
\end{minipage}
\end{table}

\subsubsection{Discussion}
Most observations and discussions in the univariate regression still hold in the multivariate task. The only difference lies in the performance of DML, where it does not improve the performance in the setting with transfer learning. The smaller size of the dataset compared to the one in the BAP task could be one possible reason. Another possible explanation is that the task's multivariate nature makes it hard to define a suitable space for DML to facilitate DKL. Further improving the DKL performance in a multivariate setting of DML would be an exciting direction for our future work. Like the BAP task, we also conducted the same experiments with the backbone of the ResNet18. More details can be found in Appendix~\ref{app: point_resnet} and \ref{app: pred_resnet}.

\section{Conclusions}
This manuscript addresses the challenge that deep neural networks (DNNs) are often unable to provide uncertainty estimates for their predictions in regression tasks. Especially in the healthcare domain, this issue could prevent the further application of DNNs.   
We propose a model that consists of a deep Convolutional Neural Network (CNN) and a sparse Gaussian Process (GP). The former part serves as a trainable feature extractor that embeds raw images into a latent space. This enables the latter part to model the similarity of all sample pairs with localized kernels more effectively in order to produce a predictive distribution for each data sample. 

We show that such an architecture can be trained in an end-to-end fashion using stochastic gradient descent (SGD). We also analyzed multiple ways to boost the performance of such a model with different initialization and pre-training methods. Our approach is by no means limited to Convolutional Neural Networks, but could be generalized to other kinds of neural networks that best fit the nature of the data. We also observe a new specific challenge in this setup: We observe that randomly initialized inducing points in a sparse GP cause the prediction to degenerate to its prior 
when it consumes outputs from CNN backbones. We propose a simple solution that could also encourage further research in the task of jointly learning representations and GPs. 
Our experiments on the Bone Age Prediction and Lesion Localization tasks show that the proposed model delivers mostly better performance in terms of point estimates if compared to the baselines with a linear output layer. More importantly, we show that our model's prediction performance increases hand-in-hand with its predictive certainty. In other words, given a difficult test sample, our model can realize and communicate that the prediction thereof might be less trustworthy by generating a larger predictive variance.  
Finally, our model requires significantly less computational cost than popular MC Dropout methods, which motivates its usage in real-world online applications.

As future work, we are interested in studying the integration of multiple sparse GPs to deal with different types of input sources and model the relations between outputs. In addition, a combination of the interpretation using our uncertainty-aware model with the explainability methods like various saliency methods\cite{gu2018understanding, gu2020introspective} would be an exciting direction for exploration.

\section*{Acknowledgment}
The authors acknowledge the support from Yan Ke on the Deep Lesion dataset\cite{yan2018deeplesion} and the support by the German Federal Ministry for Education and Research (BMBF), funding project “MLWin” (grant 01IS18050).
\begin{figure}[!htb]
    \includegraphics[width=0.5\linewidth]{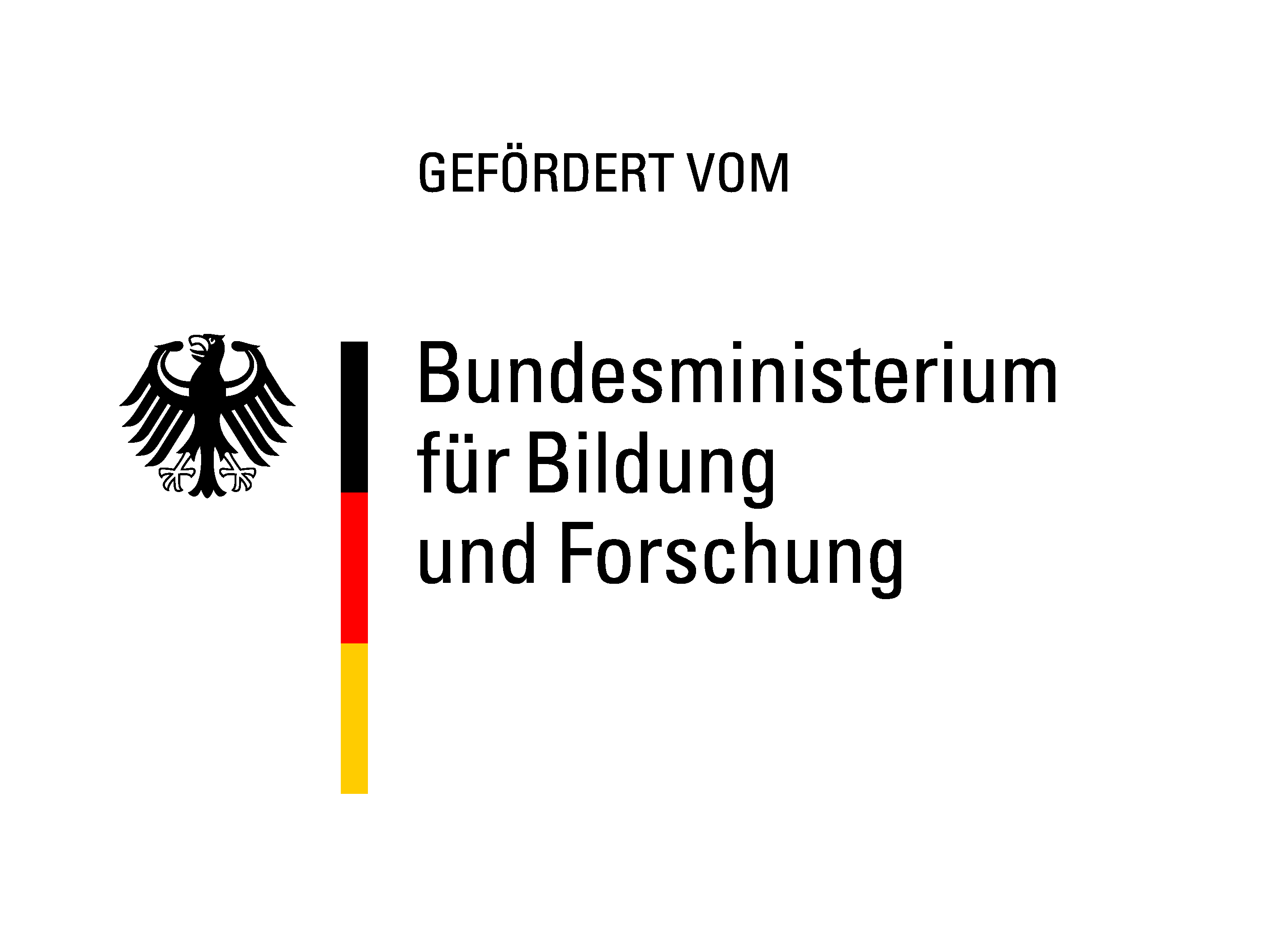}
\end{figure}
\bibliographystyle{IEEEtran}
\bibliography{Ref_ICHI2021}

\appendix

\subsection{Results on Point Estimates using ResNets}\label{app: point_resnet}

\begin{table}[!h]
\vspace{-0.2cm}
\scriptsize
\centering
\caption{Bone Age Prediction with ResNet18}
\label{tab: ba_resnet}
\begin{tabular}{ccccc}
\hline
\begin{tabular}[c]{@{}c@{}}Output \\ Layer\end{tabular} & \begin{tabular}[c]{@{}c@{}}Transfer \\ Learning\end{tabular} & \begin{tabular}[c]{@{}c@{}}RMSE\\ (No pre-training)\end{tabular} & \begin{tabular}[c]{@{}c@{}}RMSE\\ (Metric)\end{tabular} & \begin{tabular}[c]{@{}c@{}}RMSE\\ (CAE)\end{tabular} \\ \hline
Linear$^*$                                          & Yes                                                          & $13.131 \pm 0.129$                                              & $\mathbf{12.419 \pm 0.098}$                             & $13.842 \pm 0.283$                                   \\
SVGP$^\dagger$                                             & Yes                                                          & $\mathbf{12.632 \pm 0.149}$                                     & $12.567 \pm 0.051$                                      & $\mathbf{13.375 \pm 0.151}$                          \\
PPGP$^\dagger$                                         & Yes                                                          & $12.899 \pm 0.114$                                              & $12.658 \pm 0.048$                                      & $13.640 \pm 0.292$                                   \\\hline \hline
Linear$^*$                                         & No                                                           & $19.766 \pm 0.134$                                              & $16.533 \pm 0.138$                                      & $16.849 \pm 0.275$                                   \\
SVGP$^\dagger$                                        & No                                                           & $\mathbf{19.090 \pm 0.286}$                                     & $\mathbf{16.147 \pm 0.264}$                             & $16.641 \pm 0.275$                                   \\
PPGP$^\dagger$                                       & No                                                           & $20.166 \pm 0.371$                                              & $16.986\pm0.196$                                        & $\mathbf{16.469\pm0.222}$                            \\ \hline
\end{tabular}
\end{table}
\vspace{-0.4cm}
\begin{table}[h]
\begin{minipage}{0.5\textwidth}
\centering
\caption{Lesion Localization with ResNet18}
\label{tab: dl_resnet}
\begin{tabular}{ccccc}
\hline
\begin{tabular}[c]{@{}c@{}}Output \\ Layer\end{tabular} & \begin{tabular}[c]{@{}c@{}}Transfer \\ Learning\end{tabular} & \begin{tabular}[c]{@{}c@{}}RMSE\\ (No pre-training)\end{tabular} & \begin{tabular}[c]{@{}c@{}}RMSE\\ (Metric)\end{tabular} & \begin{tabular}[c]{@{}c@{}}RMSE\\ (CAE)\end{tabular} \\ \hline
Linear\footnote{Common architectures.}                   & Yes                                                          & $0.110 \pm 0.003$                                               & $0.114 \pm 0.001$                                       & $0.111 \pm 0.003$                                    \\
SVGP\footnote{With our proposed model.}             & Yes                                                          & $\mathbf{0.102 \pm 0.002}$                                      & $0.107 \pm 0.001$                                       & $0.106 \pm 0.003$                                    \\
PPGP$^\dagger$                                        & Yes                                                          & $0.103 \pm 0.001$                                               & $\mathbf{0.104 \pm 0.001}$                              & $\mathbf{0.103 \pm 0.002}$                           \\\hline \hline
Linear$^*$                                        & No                                                           & $0.148 \pm 0.003$                                               & $0.137 \pm 0.002$                                       & $0.123 \pm 0.002$                                    \\
SVGP$^\dagger$                                & No                                                           & $\mathbf{0.129 \pm 0.002}$                                      & $\mathbf{0.131 \pm 0.001}$                              & $0.121 \pm 0.002$                                    \\
PPGP$^\dagger$                                     & No                                                           & $0.133 \pm 0.002$                                               & $0.134 \pm 0.002$                                       & $\mathbf{0.119 \pm 0.002}$                           \\ \hline
\end{tabular}
\end{minipage}
\end{table}

\subsection{Results on Predictive Variances using ResNets}\label{app: pred_resnet}
\vspace{-0.2cm}
\begin{figure}[!h]
    \centering
    \includegraphics[width=0.24\textwidth]{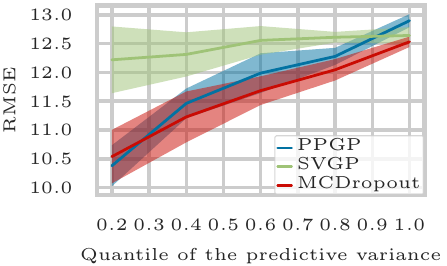}
    \includegraphics[width=0.24\textwidth]{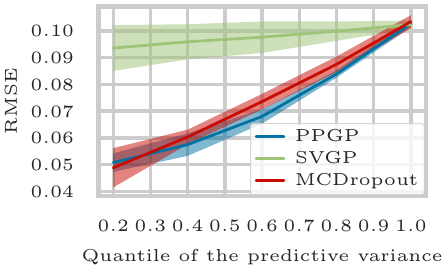}
    \caption{Quantile Performance for tasks of Bone Age Prediction (left) and Lesion Localization (right) using ResNet18 as the backbone in our model}
    \label{fig:qo_ba_resnet}
\end{figure}
    

\end{document}